\documentclass[10pt,twocolumn,letterpaper]{article}

\usepackage{times}
\usepackage{epsfig}
\usepackage{graphicx}
\usepackage{amsmath}
\usepackage{amssymb}

\renewcommand{\P}{\mathbb{P}_}

\newcommand{\E}{\mathbb{E}}
\newcommand{\R}{\mathbb{R}}

\newcommand{\pd}{\P{\text{data}}}
\newcommand{\pg}{\P{G_\text{hist}}}
\usepackage{comment}
\usepackage{color}
\usepackage[normalem]{ulem} 
\usepackage{amsfonts}
\usepackage{bbm} 

\usepackage[pagebackref=true,breaklinks=true,letterpaper=true,colorlinks,bookmarks=false]{hyperref}

\begin{document}

\title{History-based anomaly detector: an adversarial approach to anomaly detection}

\author{Pierrick CHATILLON\\
\'Ecole normale sup\'erieure Paris-Saclay\\
{\tt\small pierrick.chatillon@ens-paris-saclay.fr}
\and
Coloma BALLESTER\\
Universitat Pompeu Fabra\\
{\tt\small coloma.ballester@upf.edu}
}

\maketitle

\begin{abstract}
Anomaly detection is a difficult problem in many areas and has recently been subject to a lot of attention. Classifying unseen data as anomalous is a challenging matter. Latest proposed methods rely on Generative Adversarial Networks (GANs) to estimate the normal data distribution, and produce an anomaly score prediction for any given data. In this article, we propose a simple yet new adversarial method to tackle this problem, denoted as History-based anomaly detector (HistoryAD). It consists of a self-supervised model, trained to recognize 'normal' samples by comparing them to samples based on the training history of a previously trained GAN. Quantitative and qualitative results are presented  evaluating its performance. We also present a comparison to several state-of-the-art methods for anomaly detection showing that our proposal achieves top-tier results on several datasets.
\end{abstract}


\section{Introduction}
\label{sec:intro}

Anomaly detection usually refers to the identification of unusual patterns that do not conform to expected behaviour of data, be it visual data such as images and videos, or other modalities such as acoustics or natural language. Its applications are numerous and include the detection of anomalies in medical or biological imaging such as failure of neurocognitive functions in damaged brains~\cite{grosjean2009contrario,schlegl2017unsupervised,prokopetc2017slim}, real-life image forgery resulting in fake news or even fraud~\cite{huh2018fighting,zhou2018learning,Wu_2019_CVPR,nikoukhah2019jpeg}, anomaly detection in image or video for autonomous navigation, driver assistance systems or surveillance systems for, \emph{e.g.}, violence alerting or evidence investigation~\cite{luo2017revisit,Morais_2019_CVPR,Zhong_2019_CVPR,Nguyen_2019_ICCV,Ionescu_2019_CVPR,dwibedi2019temporal}, or for detection of violation and foul in sports analysis, detection of defective samples in manufacturing industry~\cite{tout2017automatic,zontak2010defect,Bergmann_2019_CVPR}, sea mines in side-scan sonar images~\cite{mishne2013multiscale} or extrange aerial objects in aerial images that may produce collisions~\cite{nussberger2016robust}, anomalies from multi-modal data including visual data, audio data or natural language~\cite{li2019deep}, to name but a few of its applications.
\begin{figure}[t]
\begin{minipage}[b]{1.0\linewidth}
  \centering
  \centerline{\includegraphics[width=8.5cm,height=3cm]{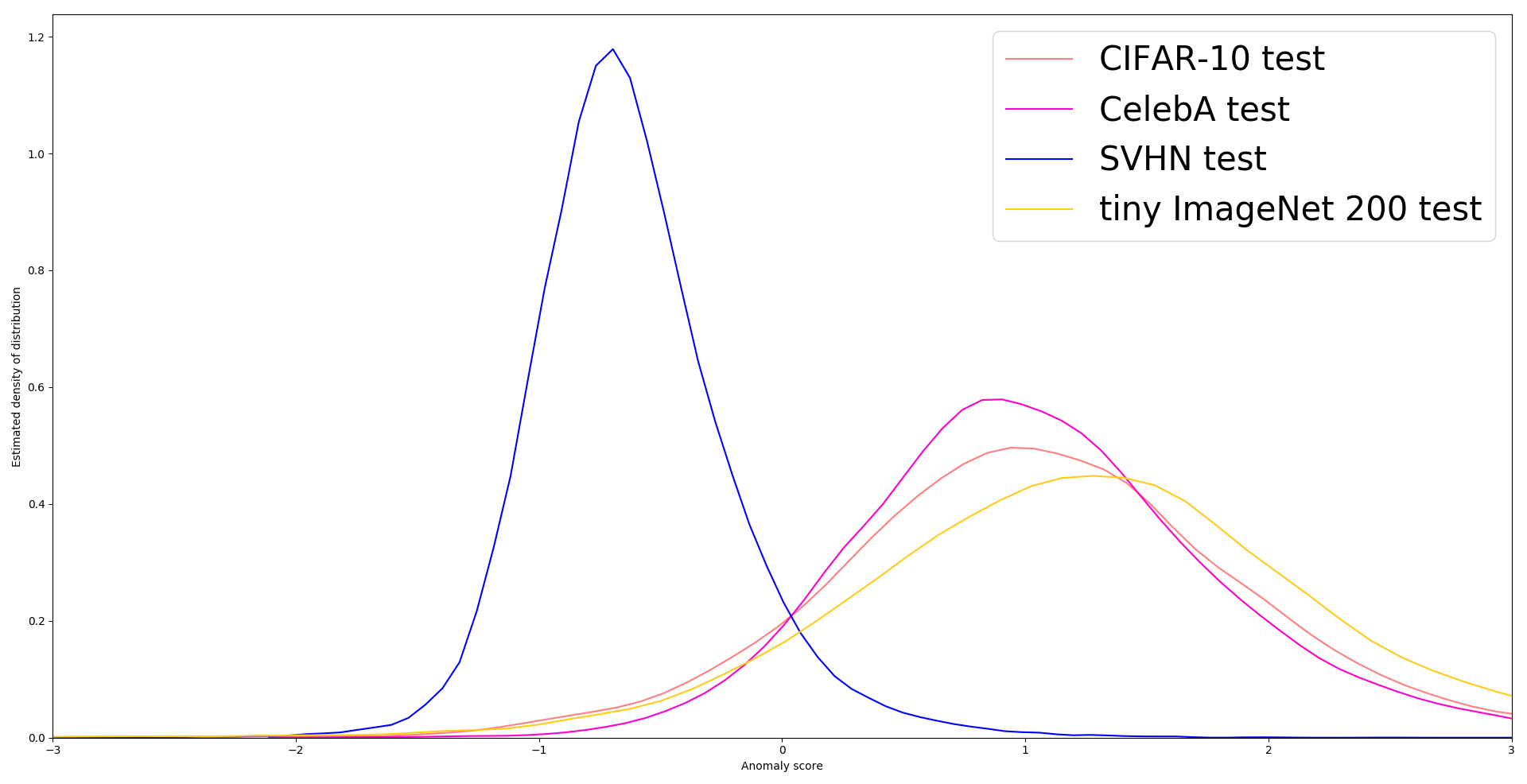}}
  \center{Approximate density of anomaly score distribution for each dataset.}\medskip
\end{minipage}

\label{svhn}
\begin{minipage}[b]{1.0\linewidth}
\begin{center}
\begin{tabular}{|l|c|c|c|}
  \hline
 test split&CIFAR-10 & CelebA  & Tiny ImageNet\\
  \hline
  AUPRC&0.941 & 0.976 & 0.949 \\
  \hline
\end{tabular}
\end{center}
\center{Table 1. AUPRC for SVHN compared to other datasets.}
\end{minipage}
\caption{Method trained on SVHN and evaluated on several datasets.}
\end{figure}

The precise definition of anomalous data is inherently difficult as, in practice, an unexpected anomaly can be detected only against the ground of a pattern regularity. This is one of the reasons that anomaly detection is frequently approached as out-of-distribution or outlier detection. A detailed account of the many existing methods to approach this problem can be found in \cite{chandola2009anomaly,pimentel2014review,ehret2019image,chalapathy2019deep}. 

This paper proposes a new method for anomaly detection in the context of image processing that is based on the unsupervised learning of the underlying probability distribution of normal data through appropriate GANs and the proposal of a new anomaly score for the detection of abnormal images. Our anomaly detector leverages a recorded history of the normal data generator to fully discriminate regions where true data points are more dense and use this learning to successfully detect anomalies. It results in a general anomaly detector that is free of assumptions on the data and thus it can be applied in any context and data modality. Fig.~\ref{svhn} illustrates an example of the performance of our anomaly detector on structurally different datasets. In this experiment, the distribution of the Street View House Numbers (SVHN) dataset \cite{SVHN} is first learned (details in Section~\ref{sec:method}) and considered as normal data. Then, our anomaly score is computed on samples of it and also on samples from the CIFAR-10~\cite{krizhevsky2009learning}, CelebA \cite{CelebA} and  ImageNet \cite{imagenet_cvpr09} datasets. The approximated density of the anomaly score distribution for each dataset is shown at the top of Fig.~\ref{svhn}. Let us notice that, for the normal data, its anomaly values are around $-1$ while for all the 'anomalous' datasets, the anomaly scores are around $+1$. On the other hand, Table 1 in Fig.~\ref{svhn} bottom shows the area under the precision-recall curve (AUPRC).

The outline of this paper is as follows. Section~\ref{sec:relatedwork} reviews related research. Section~\ref{sec:method} details our proposal. In Section~\ref{sec:algo}, the model architecture and implementation details are provided while the experimental results are presented in Section~\ref{sec:results}. Finally, the paper is concluded in Section~\ref{sec:conclusions}.

\section{Related Work}
\label{sec:relatedwork}
The automatic identification of abnormal or manipulated data is crucial in many contexts~\cite{schlegl2017unsupervised,schweizer2000hyperspectral,huh2018fighting,Wu_2019_CVPR,nikoukhah2019jpeg,Zhong_2019_CVPR,Bergmann_2019_CVPR}. Anomaly detection, has been a topic in statistics for centuries (see \cite{chandola2009anomaly,pimentel2014review,ehret2019image,chalapathy2019deep} and references therein). The authors of~\cite{ehret2019image} classify the methods in the literature by the structural assumption made on the normality. 
Other works challenge anomaly detection 
with unsupervised or self-supervised learning strategies by taking advantage of the huge amount of data frequently at our disposal~\cite{an2015variational,SchleglSWSL17}. 
Some of them use generative models to learn the (normal) data distribution. Generative models are methods that produce novel samples from high-dimensional data distributions, such as
images or videos. 
Currently the most prominent approaches include autoregressive models \cite{van2016conditional}, variational autoencoders (VAE) \cite{vae}, and generative adversarial networks (GANs) \cite{GAN}. GANs are often credited for producing less burry outputs when used for image generation. In the anomaly detection context, several approaches tackle it using autoencoders~\cite{Gong_2019_ICCV} or GANs \cite{SchleglSWSL17,zenati2018efficient,deecke2018image,ravanbakhsh2017abnormal,haloui2018anomaly,akcay2018ganomaly,ssod,DBLP:journals/corr/abs-1904-01209} (we refer to \cite{DBLP:journals/corr/abs-1906-11632} for a summary of those GAN-based anomaly detection methods).
Some works focus on the implicit inversion of the generator in order to detect anomalous data that do not fall in the learned model \cite{DBLP:journals/corr/DonahueKD16,haloui2018anomaly}, while others directly infer likelihoods with, for instance, normalizing flows \cite{kingma2018glow,choi2018waic,hendrycks2018deep,nalisnick2018deep,serra2019input}. On the other hand, the recent paper~\cite{Gong_2019_ICCV} uses a memory-augmented autoencoder which learns and records a fixed number of prototypical normal encoded vectors. Given an input sample, it is encoded and the memory is then accessed with an attention-based module to express this encoding by a sparse combination of the stored normal prototypes that used to reconstruct the input data via a decoder. The $l_2$ distance between the input and its reconstruction 
is used as anomaly score. Very differently, our anomaly detector leverages the recorded history of the normal data generator to fully discriminate regions where true data points are more dense and use this learning to successfully detect anomalies. 
The idea of producing an anomaly score prediction for any given data has also been investigated~\cite{haloui2018anomaly,SchleglSWSL17,akcay2018ganomaly}. Our proposal fits in the class of self-supervised approaches and it is trained only on normal (non-anomalous) samples. The proposed method is general, efficient and simple as it uses the rich information of the training process in the construction the anomaly detector.

\section{Proposed method}

\label{sec:method}
We will attribute an anomaly score to any image. Inspired by some ideas in \cite{ssod} and \cite{DBLP:journals/corr/abs-1904-01209}, this score consists of the output of a network.

\begin{figure*}
\begin{center}
\centerline{\includegraphics[width=\linewidth]{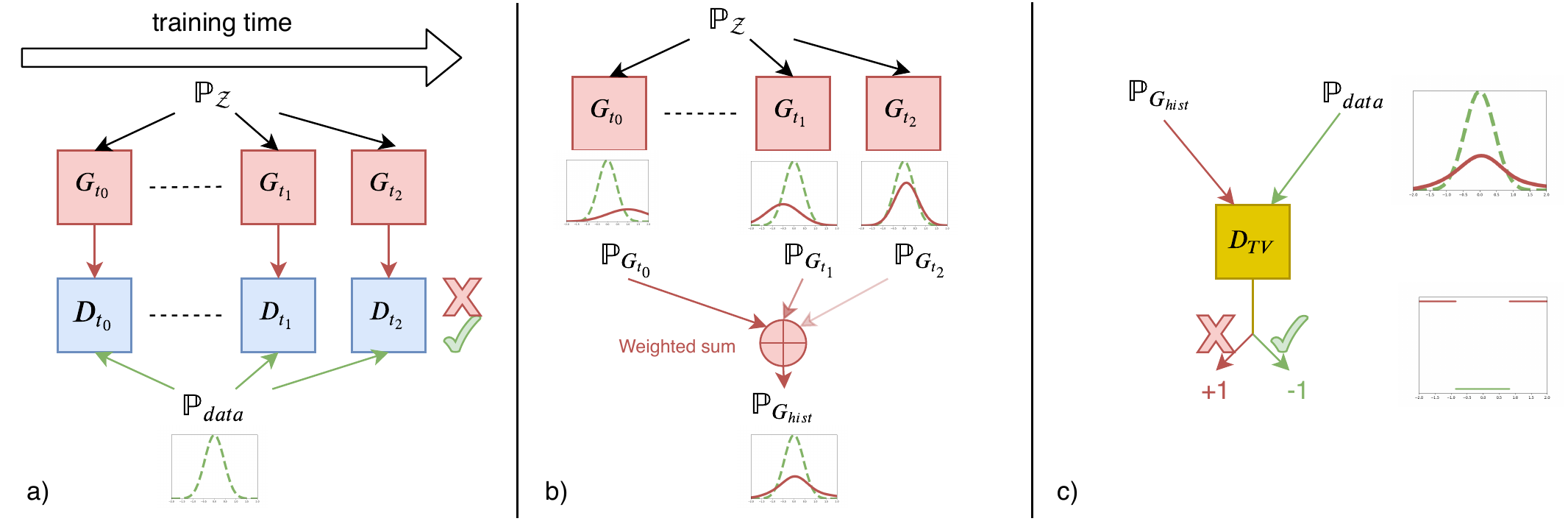}}
\end{center}

\caption{\textbf{Outline of the proposed method:} a) Some states of the generator are saved during GAN training ($G_{t_{i}}$ and $D_{t_{i}}$ represent the states of of the generator and discriminator at training time $t_i$ ). b) these networks are used to form a new distribution $\P{G_{\text{hist}}}$ rich in 'anomalous' samples. c) We use this distribution as negative class for a classifier. $D_{TV}$}\label{method}
\end{figure*}

More precisely, let $\P{\text{data}}$ be the probability distribution of a given 'normal images' dataset. Our proposal is grounded on, first, the learning of the probability distribution $\P{\text{data}}$ using a GAN learning strategy while simultaneously keeping track of the states of the associated generator and discriminator during training. Secondly, we create a probability distribution (denoted as $\P{G_\text{hist}}$) that combines different states of the previous generator's history. We finally train our anomaly detector by computing the Total Variation distance between the real data distribution $\P{\text{data}}$ and $\P{G_\text{hist}}$. 
Fig. \ref{method} displays an outline of the whole method, and
in the following Sections~\ref{sec:first}, \ref{sec:second}, and \ref{sec:third}, these steps of our approach are detailed and justified.

\subsection{Learning to generate training-like data}\label{sec:first}
As mentioned, a GAN-based adversarial strategy is followed. Let us recall that the GAN strategy \cite{GAN} is based on a game theory scenario between two networks, the generator and the discriminator, having adversarial objectives. The generator maps a noise vector (of density $\P {Z}$) from the latent space to the image space trying to trick the discriminator, while the discriminator receives either a generated or a real image and must distinguish between both. This procedure leads the probability distribution of the generated data to be as close as possible, for some distance, to the one of the real data. For the Vanilla GAN \cite{GAN}, the minimized distance is the Jensen-Shannon Divergence, which has arguably bad properties (see Section 2 of \cite{arjovsky2017wasserstein} for details).
The authors of \cite{arjovsky2017wasserstein} introduced the idea of minimizing the Wasserstein-1 distance (denoted as $\mathbb{W}_1$) instead. They proved several of its nice properties, including its continuity and differentiability almost everywhere (under certain hypotheses). The Wasserstein-1 distance can be computed with the Kantorovich duality property: if $\P1$ and $\P2$ are two probability distributions, then
\begin{equation}\label{eq:Wass}
\mathbb{W}_1(\P1,\P2)=\sup_{D\in{\cal{D}}} \E_{x\sim\P1}[D(x)]- \E_{y\sim\P2}[D(y)], 
\end{equation}
where ${\cal{D}}$ is set of 1-Lipschitz functions, i.e., in the notations of \cite{villani2008optimal}, the set of c-convex functions for the cost function $c(x, y)=\mid x-y \mid$.
Let $G$ and $D$ be the generator and the discriminator learned by optimizing the adversarial Wasserstein GAN loss (WGAN),
\begin{equation}\label{eq:WGAN}
\inf_G \sup_{D\in{\cal D}} \, {\mathbb{E}}_{x\sim \P{G}}\left[ D(x)\right] - {\mathbb{E}}_{x\sim \pd}\left[ D(x)\right] .
\end{equation}
Notice that the optimal dual variable $D^*$ obtained from the optimization of (\ref{eq:WGAN}) will be negative on real data samples and positive on generated ones. In this paper, we use the learning strategy of \cite{WGAN-GP} which is based on approximating the class $\mathcal{D}$ by neural networks D subject to a gradient penalty (forcing the $L^2$ norm of the gradient of the discriminator with respect to its input to be close to 1).
The choice of WGAN instead of other GAN losses favours nice properties such as avoiding vanishing gradients and mode collapse, and achieves more stable training.
\begin{figure}[t!]
\begin{minipage}[t]{.49\linewidth}
  \centering
  \centerline{\includegraphics[width=4cm,height=3cm]{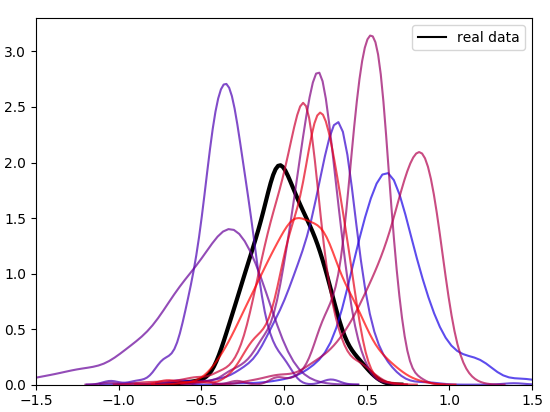}}
  \center{(a) Empirical generated distributions; color from blue to red indicates the progression of training.}
\end{minipage}
\begin{minipage}[t]{.49\linewidth}
  \centering
  \centerline{\includegraphics[width=4cm,height=3cm]{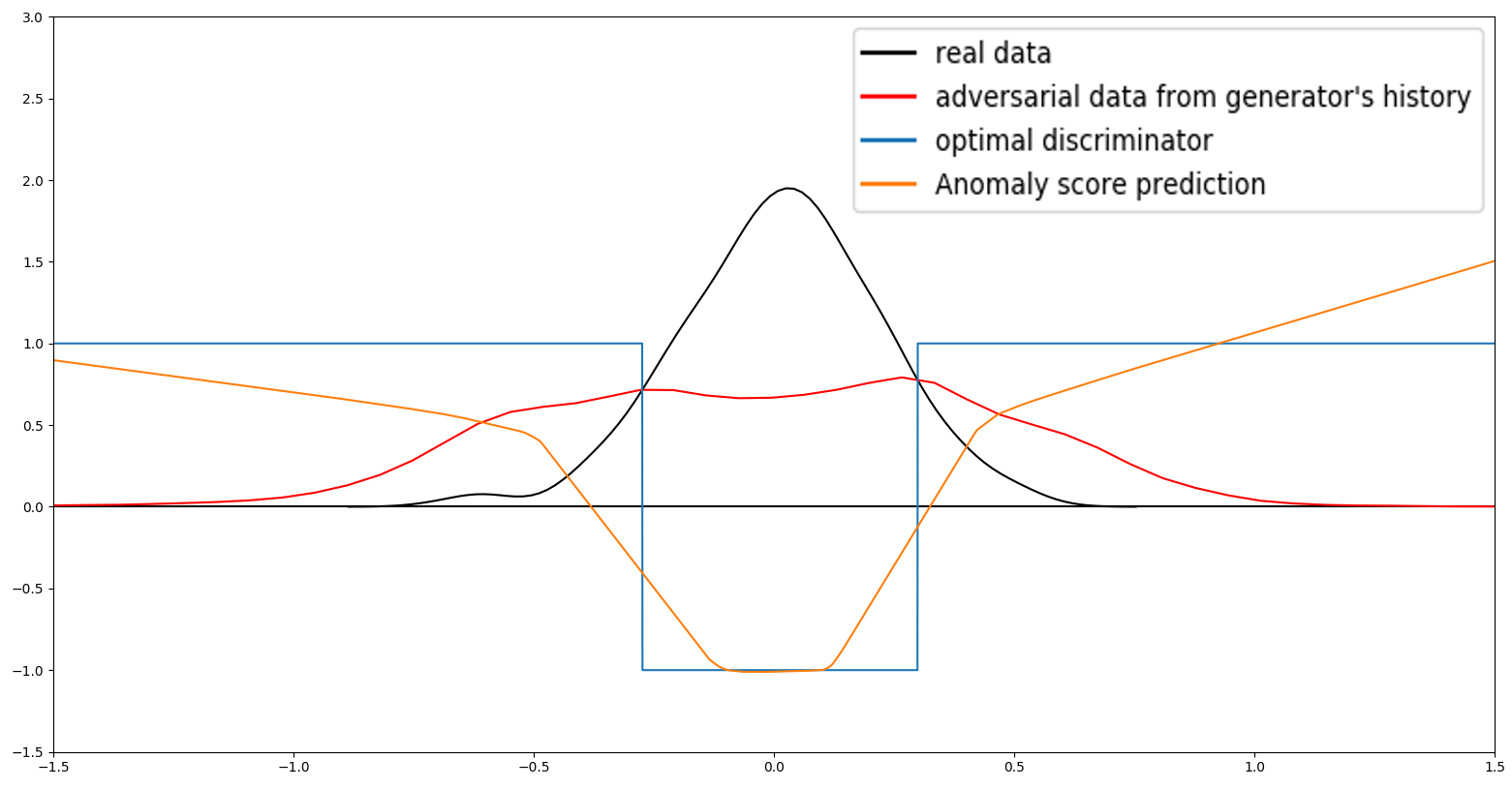}}
  \center{(b) Comparison of $\P{\text{data}}$ and $\P{G_\text{hist}}$ and profile of optimal $D_{TV}^{*}$ and trained $D_{TV}$.}\medskip
\end{minipage}
\caption{Method illustration on a toy one-dimensional dataset of points sampled from a normal law.}\label{1d_ex}
\end{figure}

\subsection{Generator's history probability distribution}\label{sec:second}
Let us start by presenting the underlying idea. During training (equation (\ref{eq:WGAN}) with algorithm of \cite{WGAN-GP}), the discriminator $D$ will indicate regions that may contain real data, and $G$ learns to produce samples in that zone.
If these zones do not contain real data, then the discriminator will act as a critic and indicate it to the generator and point at other regions.
This way, screenshots of the generator during training keep track of data points surrounding the real data manifold.
In this paper, we merge the screenshots of the generator during training to form:
\begin{equation}\label{eq:Ghist}
    \P{G_\text{hist}} \triangleq \int_{\alpha}^{n_{\textrm{epochs}}} c \cdot G_t(\P {Z}) \cdot e^{-\beta t} dt,
\end{equation}
(see Fig. \ref{method}) where $G_t$ denotes the state of the generator at training time $t$ and $\P {Z}$ is the latent space distribution (parameters $\alpha$ and $\beta$ are discussed in Section \ref{sec:algo}).
As a weighted mean of training generated distributions, we may assume by construction that $\P{G_\text{hist}}$ covers $\P{\text{data}}$. That is,
\begin{equation}\label{eq:H}
\textrm{\textbf{Hypothesis}:}\quad \textrm{supp}\left(\P{\text{data}}\right) \subset \textrm{supp}\left(\P{G_\text{hist}}\right). \qquad\quad\quad
\end{equation}
To illustrate our hypothesis and our whole method, we present a proof of concept by creating a toy one-dimensional dataset of points sampled from the normal law. 
We then train the WGAN, with the generator initialized with an offset so that it does not match training data.
As previously explained (details in Section~\ref{sec:algo}) we save the states of the generator. Fig.~\ref{1d_ex}(a) displays empirical generated distributions of some of these states.

In order to satisfy Hypothesis (\ref{eq:H}), we use momentum based optimizers, so that $\P{G}$ oscillates around $\P{\text{data}}$ (see Fig.~\ref{1d_ex}(a)), making the support of $\P{G_{\text{hist}}}$ cover the one of $\P{\text{data}}$ better (see Fig.~\ref{1d_ex}(b), where $\P{\text{data}}$ and $\P{G_{\text{hist}}}$ are, respectively, plotted in black and red).

This empirically confirms the hypothesis in this toy case.

\begin{figure}[t!]
\begin{minipage}[t]{.49\linewidth}
  \centering
  \centerline{\includegraphics[width=4cm,height=2cm]{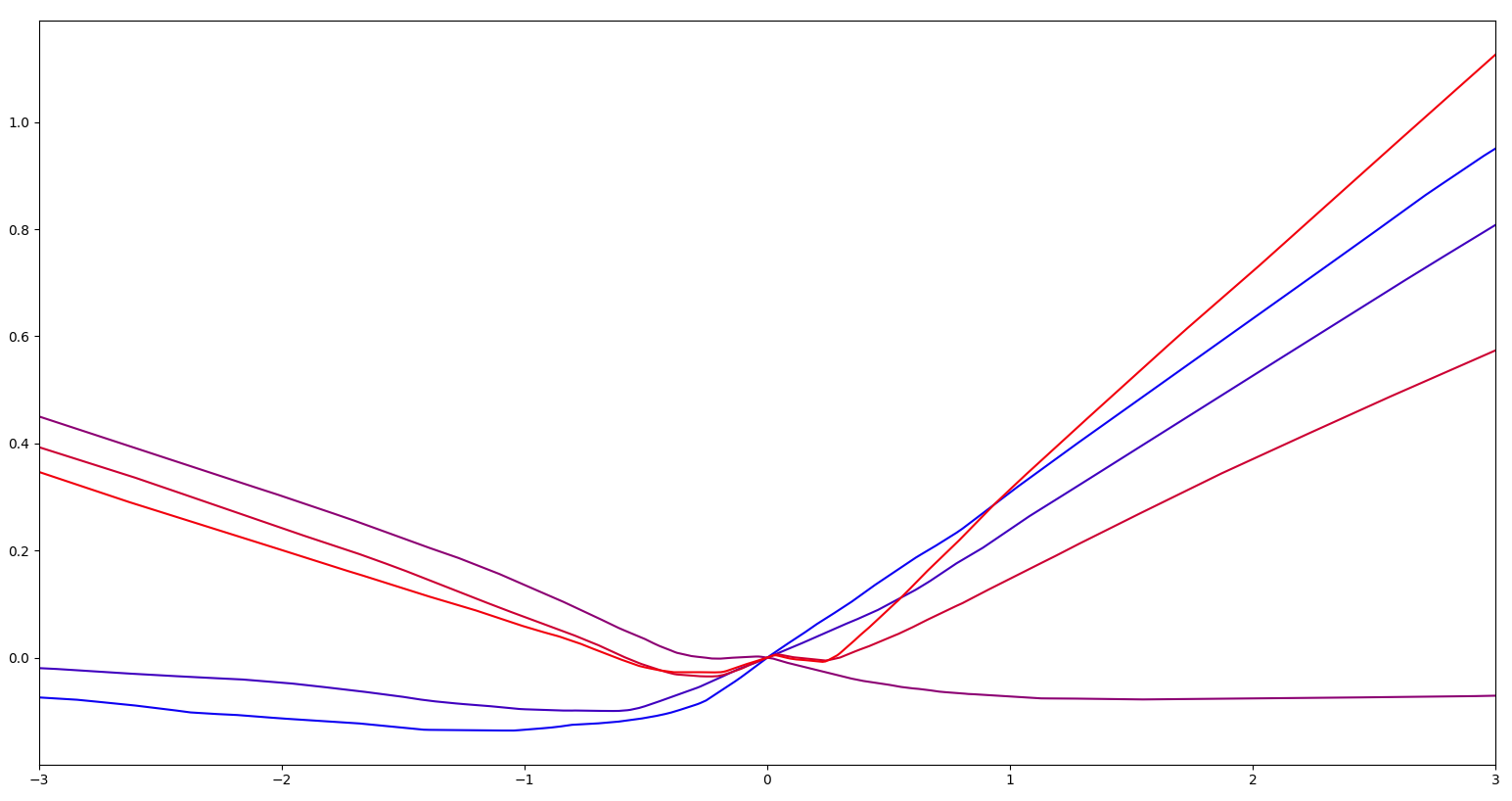}}
  \center{(a) Discriminator score output during training (from blue to red).}
\end{minipage}
\hfill
\begin{minipage}[t]{0.49\linewidth}
  \centering
  \centerline{\includegraphics[width=4cm,height=2cm]{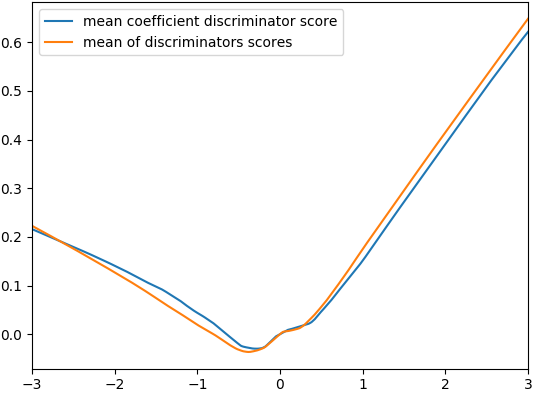}}
  \center{(b) Average outputs of discriminators versus output of a average coefficient discriminator.}\medskip
\end{minipage}
\caption{Justification of $D_{TV}$ initialization on the toy example. }
\label{1d_disc}
\end{figure}

\subsection{Training our anomaly detector $D_{TV}$}\label{sec:third}
As announced above, we will compute the Total Variation distance between $\P{\text{data}}$  and  $\P{G_\text{hist}}$. Let us explain how and why.
Firstly, we recall the Total Variation distance definition:
\begin{equation}\label{eq:TV}
\delta(\P1,\P2)=\sup_{A \text{ Borel subset} }\left|\mathbb{P}_{1}(A)-\mathbb{P}_{2}(A)\right|,
\end{equation}
which represents the choice $c(x, y)=\mathbbm{1}_{x \neq y}$ in the optimal transport problem, as stated in \cite{villani2008optimal} (where $\mathbbm{1}_{A}$ denotes the indicator function of a set $A$, as usual). As noticed by several authors (see, \emph{e.g.},~\cite{arjovsky2017wasserstein}), the topology induced by the Total Variation distance is stronger that the one induced by the Wasserstein-1 distance.
Let us remark that $\ \delta(\P1,\P2)=\frac{1}{2}\|\P1 - \P2\|_{TV}$, 
where $\|\cdot\|_{TV}$ denotes the Total Variation norm. 
The Kantorovich duality yields:
\begin{equation}\label{eq_tv} 
2\ \delta(\P1,\P2)=\sup_{-1\leq D \leq 1} \left( \E_{x\sim\P1}[D(x)]- \E_{y\sim\P2}[D(y)] \right).
\end{equation}
From this equation (\ref{eq_tv}), we infer our ideal training objective:
\begin{equation}\label{eq:obj}
\sup_{-1\leq D \leq 1} \,  {\mathbb{E}}_{x\sim \pg}\left[ D(x)\right] - {\mathbb{E}}_{x\sim \pd}\left[ D(x)\right].
\end{equation}
Let us notice that the optimal state $D^{*}$ in (\ref{eq_tv}) is completely understood:
Paraphrasing  \cite{arjovsky2017wasserstein}, take $\mu=\P1-\P2$, which is a signed
measure, and $(P, N)$ its Hahn decomposition ($P=\{d\P1 > d\P2\})$. Then, we can define $D^{*} :=\mathbbm{1}_{P}-\mathbbm{1}_{N}$, we have $-1\leq D^{*} \leq 1$, and
\begin{equation}
    \begin{aligned} E_{x \sim \mathbb{P}_{1}}\left[D^{*}(x)\right]-\mathbb{E}_{x \sim \mathbb{P}_{2}}\left[D^{*}(x)\right] &=\int D^{*} \mathrm{d} \mu\\&
=\mu(P)-\mu(N) \\&=\|\mu\|_{T V} \\  &=2\ \delta\left(\mathbb{P}_{1}, \mathbb{P}_{2}\right) \end{aligned}
\end{equation}
which closes the duality gap with the Kantorovitch optimal transport primal problem, hence the optimality of the dual variable $D^{*}$.

Now, we can approximate the Total Variation distance between $\P{\text{data}}$ and $\P{G_\text{hist}}$ by optimizing (\ref{eq:obj}) over $D_{TV}$, our neural network approximation of the dual variable $D$.

Several authors (\emph{e.g.}, 
\cite{deecke2018image}) have pointed out that the output of a discriminator obtained in the framework of adversarial training is not fitted for anomaly detection. Nevertheless, notice that our discriminator $D_{TV}$ deals with two fixed distributions, $\P{\text{data}}$ and $\P{G_\text{hist}}$. Here, the purpose of computing Total Variation distance is only used to reach the  optimal $D_{TV}^{*}$ in this well-posed problem, assuming Hypothesis (\ref{eq:H}).
$D_{TV}$ should converge to $D_{TV}^{*}=\mathbbm{1}_{P}-\mathbbm{1}_{N}$ where $(P, N)$ is the Hahn decomposition of $d\P{G_\text{hist}}-d\P{\text{data}}$ (see Fig.~\ref{1d_ex}(b), blue curve).
Importantly, we hope that thanks to the structure of the data ($\P{G_\text{hist}}$ covering $\P{\text{data}}$), $D_{TV}$ will be able to generalize high anomaly scores on unseen data. Again, this seems to hold true in our simple case: The orange curve in Fig.~\ref{1d_ex} keeps increasing outside of $\text{supp}(\P{G_\text{hist}})$.

To avoid vanishing gradient issues, we enforce the 'boundedness' condition on $D_{TV}$ not by a $tanh$ non-linearity (for instance), but by applying a smooth loss (weighted by $\lambda>0$) to its output:
\begin{equation}\label{eq:res}
     \lambda \cdot d(D_{TV}(x),[-1,1])^2
\end{equation}
where $d(v,[-1,1])$ denotes the distance of a real value $v\in\R$ to the set $[-1,1]$.

Our final training loss, to be minimized, reads:
\begin{align}
\mathcal{L}(D)= &{\mathbb{E}}_{x\sim \pd}\left[ D(x)\right] - {\mathbb{E}}_{x\sim \pg}\left[ D(x)\right]  \label{mixed objective} \\&+ \lambda \, {\mathbb{E}}_{x\sim \frac{\pd+\pg}{2}} [d(D(x),[-1,1])^2]\nonumber
\end{align}
As proved in the Appendix, the optimal D for this problem is $D^*=D_{TV}^*+\Delta^*$, with
\begin{equation}
    \Delta^*(x)=\frac{d\pg(x)-d\pd(x)}{\lambda(d\pg(x)+d\pd(x))}
\end{equation}
and the minimum loss is
\begin{equation}\label{eq:minloss}
    -2\cdot \delta(\pd,\pg) -\frac{1}{2\lambda}\int \frac{(d\pg(x)-d\pd(x))^2}{(d\pg(x)+d\pd(x))} dx.
\end{equation}
By letting $\lambda \longrightarrow \infty$, the second term in (\ref{mixed objective}) becomes an infinite well regularization term which enforces $-1\leq D \leq 1$, approaching the solution of (\ref{eq:obj}). This explains why the second term in (\ref{eq:minloss}) vanishes when $\lambda \longrightarrow \infty$. In practice, for better results, we allow a small trade-off between the two objectives with $\lambda = 10$.

From now on, we will use the output of $D_{TV}$ as anomaly score.
In a nutshell, our method should fully discriminate regions where data points are more dense than synthetic anomalous points from $\P{G_\text{hist}}$.
This yields a fast feed-forward anomaly detector that ideally assigns $-1$ to normal data and $1$ to anomalous data. Fig.~\ref{svhn} shows an example.

Our method also has the advantage of staying true to the training objective, not modifying it as in \cite{DBLP:journals/corr/abs-1904-01209}. Indeed, the authors of \cite{DBLP:journals/corr/abs-1904-01209} implement a similar method but using a non-converged state of the  generator as an anomaly generator. In order to achieve this, they add a term to the log loss that prevents the model from converging all the way.


\section{Model architecture and implementation details}\label{sec:algo}

In this section, the architecture and implementation of each of the three steps detailed in Section~\ref{sec:method} is described.

\subsection{Architecture description.}
$G$ uses transpose convolution to upscale the random features, Leaky ReLU non-linearities and BatchNorm layers.
$D$ is a classic Convolutionnal Neural Network for classification, that uses pooling downscaling, Leaky ReLU before passing the obtained features through fully connected layers.
Both G and D roughly have 5M parameters.
$D_{TV}$ has the same architecture as $D$.

\subsection{Learning to generate training-like data.}
We first train until convergence of $G$ and $D$ according to the WGAN-GP objective of Section~\ref{sec:first} for a total of $n_{\textrm{epochs}}$ epochs, and save the network states at regular intervals (50 times per epoch).
We optimize our objective loss using Adam optimizers, with decreasing learning rate initially equal to $5 \cdot 10^{-4}$.

\subsection{Generator's history probability distribution}
As announced in the previous section, if the training process were to be continuous we would arbitrarily define $\P{G_\text{hist}}$ by (\ref{eq:Ghist}), that is,   $\P{G_\text{hist}} = \int_{\alpha}^{n_{\textrm{epochs}}} c \cdot G_t(\P {Z}) \cdot e^{-\beta t} dt$, where $c$ is a normalization constant that makes $\P{G_\text{hist}}$ sum to 1. We avoid the first $\alpha$ epochs to avoid heavily biasing $\P{G_\text{hist}}$ in favour of the initial random state of the generator. The exponential decay gives less importance to higher fidelity samples at the end of the training.
In practice, we approximate $\P{G_\text{hist}}$ by sampling data from $\P{G_t}$ where $t$ is a random variable of density of probability: 
\begin{equation}\label{eq:Pgh}
c \cdot \mathbbm{1}_{[\alpha,n_{\textrm{epochs}}]}  \cdot e^{-\beta t}
\end{equation}

\begin{figure}[!t]
\begin{minipage}[b]{1.0\linewidth}
  \centering
  \centerline{\includegraphics[width=\linewidth,height=5cm]{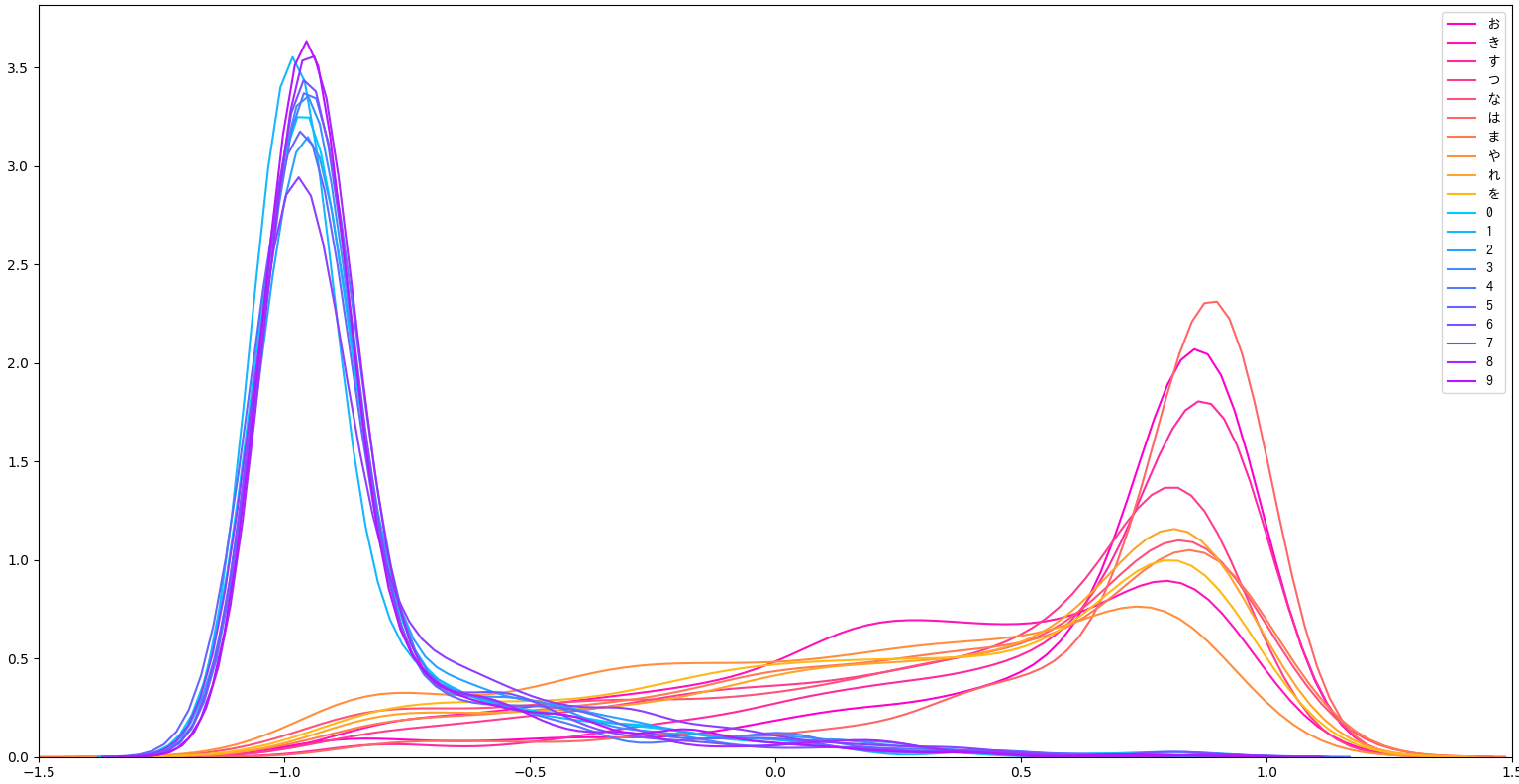}}
  \center{Approximate density of score distribution among each class of datasets (blue shades: MNIST, red shades: KMNIST).}\medskip
  
\end{minipage}
\begin{minipage}[b]{.48\linewidth}
  \centering
  \centerline{\includegraphics[width=4.5cm,height=3cm]{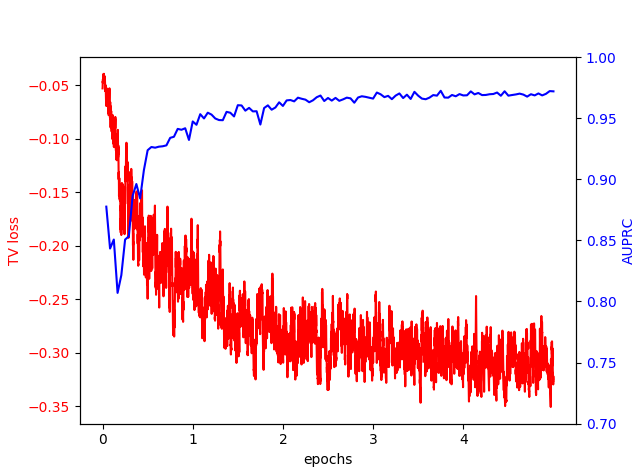}}
  \center{Red curve: Total Variation distance minimisation. Blue curve: AUPRC of the model during training.}\medskip
\end{minipage}
\hfill
\begin{minipage}[b]{0.48\linewidth}
  \centering
  \centerline{\includegraphics[width=4.5cm,height=3cm]{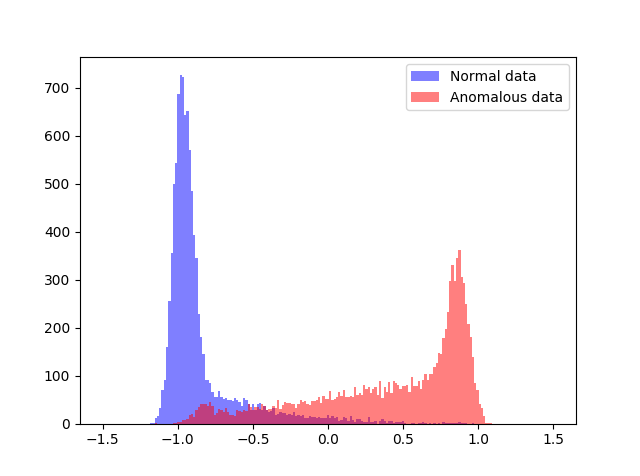}}
  \center{Histograms of anomaly scores for the test splits of MNIST and KMNIST.}\medskip
\end{minipage}

\begin{minipage}[b]{1.0\linewidth}
  \centering
  \centerline{\includegraphics[width=8cm,height=1.6cm]{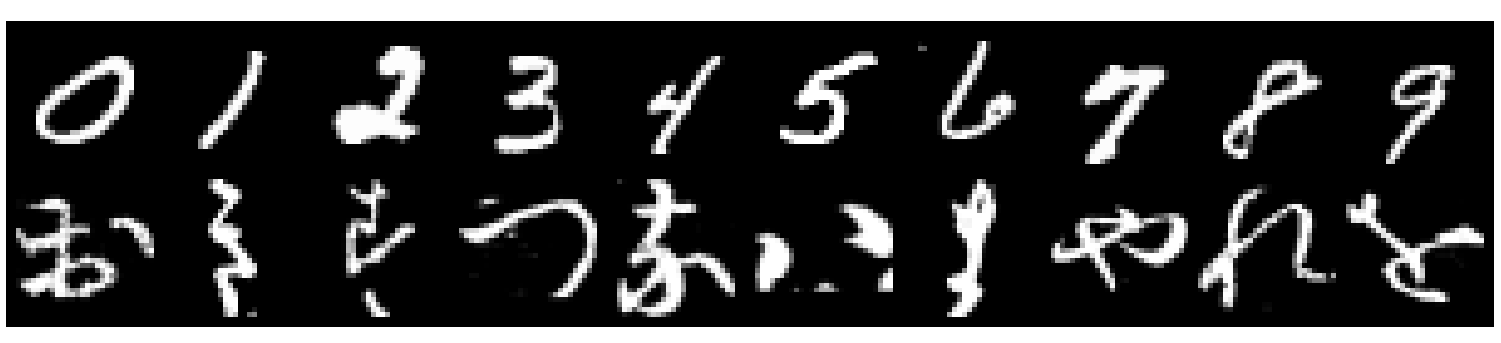}}
  \center{One sample from each class of MNIST and KMNIST}.
\end{minipage}
\caption{Method trained on MNIST (normal) and evaluated on KMNIST (anomalous).}
\label{mnistkmnist}
\end{figure}

\subsection{Training our anomaly detector $\mathbf{D_{TV}}$.}

\begin{figure*}
\begin{center}
\includegraphics[width=\linewidth,height=7cm]{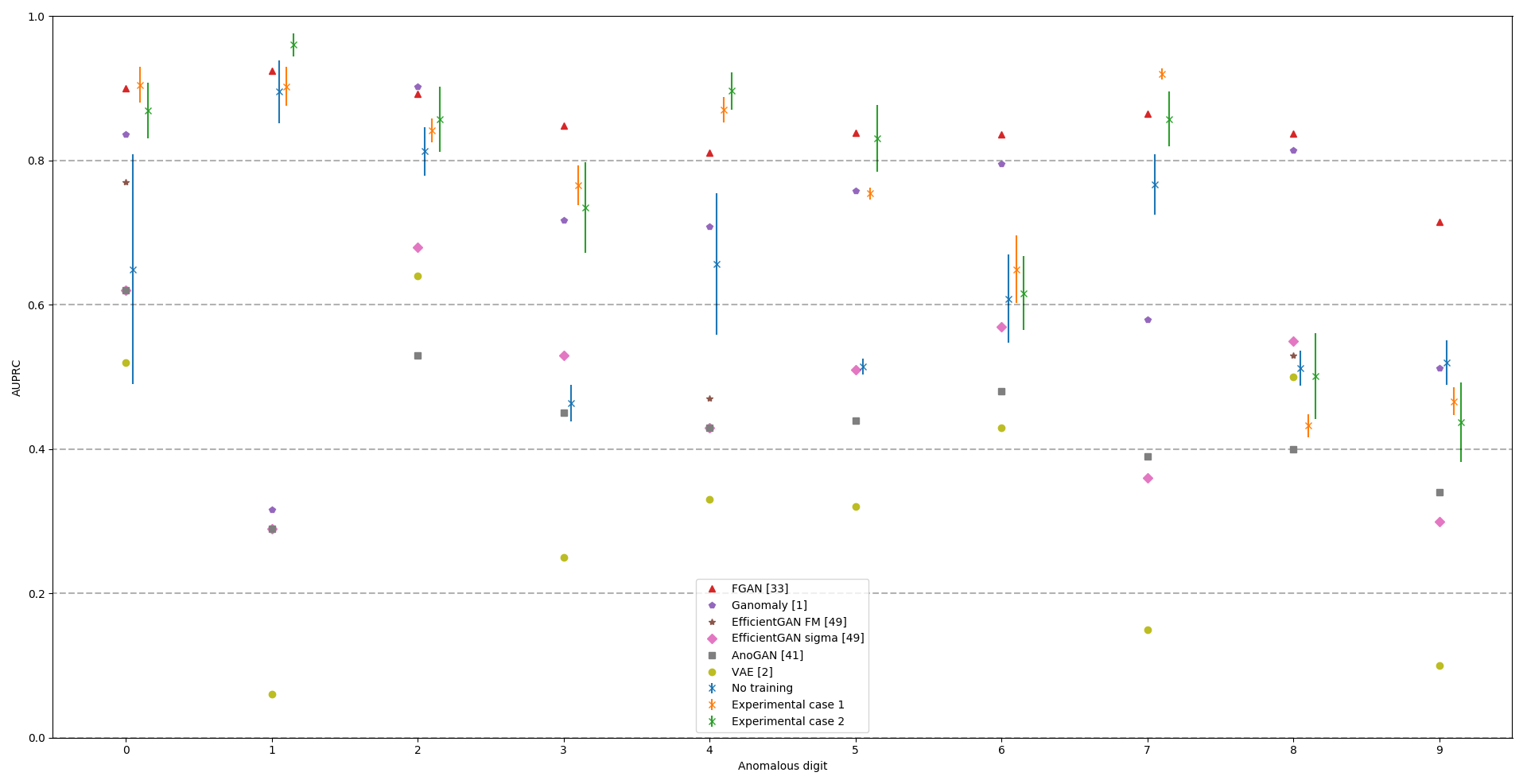}
\end{center}
   \caption{Comparison of AUPRC with other methods (x-axis denotes the MNIST digit chosen as anomalous).}
\label{fig:scores}
\end{figure*}

%
$D_{TV}$ is also optimized using Adam algorithm. Since it has the same architecture as the discriminator used in the previous WGAN training, its weights, denoted as $W_{D_{TV}}$, can be initialized as:
\begin{equation}
    W_{D_{TV}}=\int_{\alpha}^{n_{\textrm{epochs}}} c \cdot W_{D_t} \cdot e^{-\beta t} dt,
\end{equation}
where $D_t$ is the state of the WGAN discriminator at training time t.
Let us comment on the reason of such initialization.
Fig.~\ref{1d_disc}(a) seems to indicate that averaging the discriminators' outputs is a good initialization. The obtained average is indeed shaped as a 'v' centered on the real distribution in our toy example.
Fig.~\ref{1d_disc}(b) empirically shows that the initialization of the discriminator with average coefficients is somewhat close to the average of said discriminators.
As explored in \cite{DBLP:journals/corr/abs-1811-10515}, this Deep Network Interpolation is justified by the strong correlation of the different states of a network during training. 
To further discuss how good is exactly this initialization, Fig.~\ref{fig:scores} (blue error bars in the figure) shows a comparison of area under the precision-recall curve (AUPRC) with other methods on the MNIST dataset. The x-axis indicates the MNIST digit chosen as anomalous.

The following experimental cases are tested:
\begin{itemize}
    \item \textbf{Experimental case 1:} The training explained above is implemented.
    
\item \textbf{Experimental case 2:}  The same process is applied, only modifying $\P{G_\text{hist}}$, corrupting half generated images, by sampling the latent variable with a wider distribution $\P {Z'}$:\newline $\P{G_\text{hist}} =  \int_{\alpha}^{n_{\textrm{epochs}}} c \cdot \frac{G_t(\P {Z})+G_t(\P {Z'})}{2} \cdot e^{-\beta t} dt$.\\
The idea behind this is encouraging $\P{G_\text{hist}}$ to spread its mass further away from $\P{\text{data}}$.
\end{itemize}
    
\begin{figure}[!t]
\begin{minipage}{1.0\linewidth}
  \centering
  \centerline{\includegraphics[width=8.5cm,height=2.5cm]{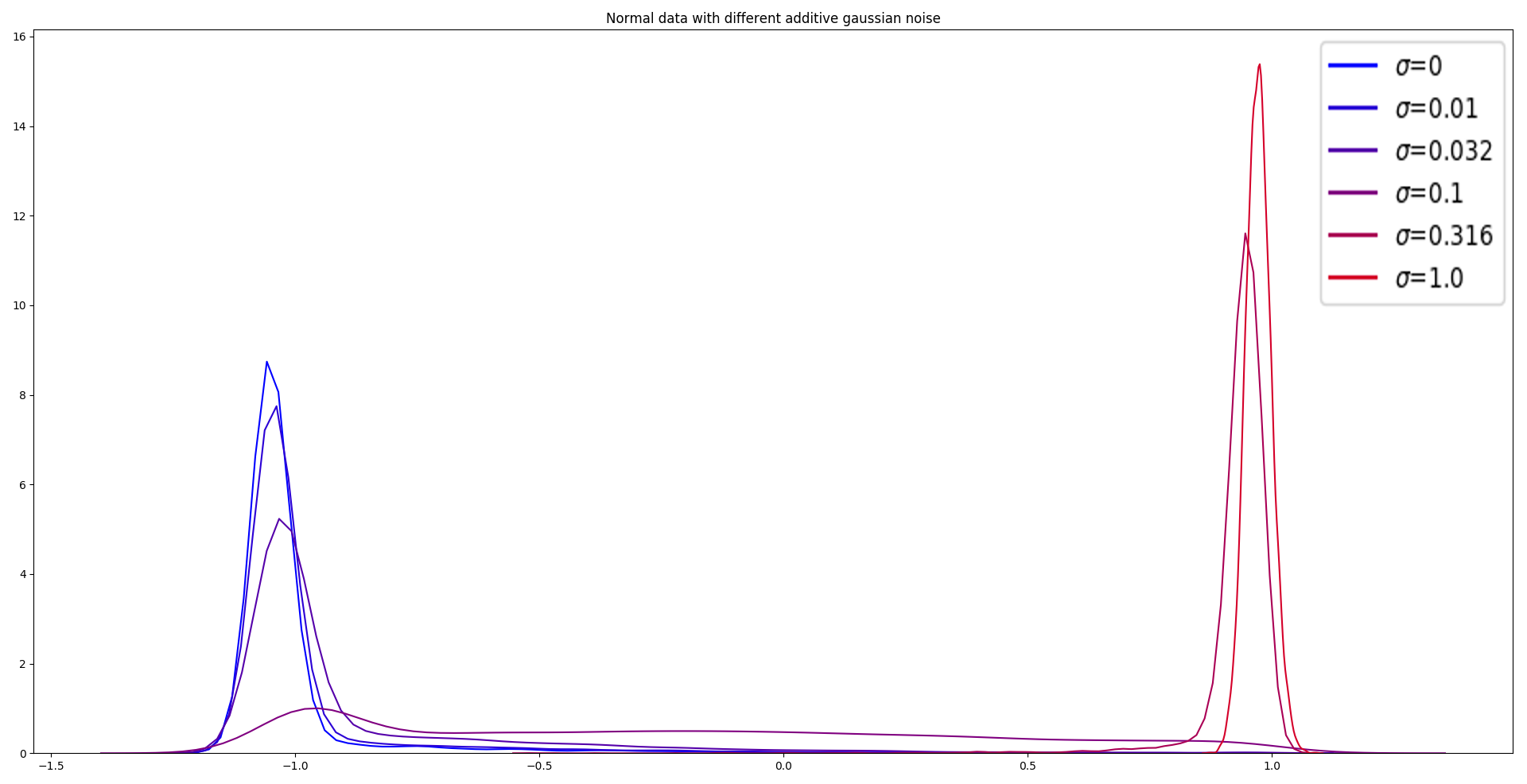}}
\end{minipage}
\caption{Method trained on MNIST (normal) and evaluated on modified MNIST images for different levels $\sigma$ of gaussian additive noise.}
\label{mnist/noise}
\end{figure}

Fig.~\ref{svhn} was computed on experimental case 1 with $n_{\textrm{epochs}}=10$, $\alpha=1$ and $\beta=5$.  Fig.~\ref{mnistkmnist}, \ref{mnist/noise} and \ref{fig:scores} were computed with $n_{\textrm{epochs}}=5$, $\alpha=1$, $\beta=3$. 

\section{Experimental Results and Discussion}
\label{sec:results}
This section presents quantitative and qualitative experimental results.

Our model behaves as one would expect when presented normal images modified with increasing levels of noise, which is attributing an increasing anomaly score to them. This is illustrated in Fig. \ref{mnist/noise} where a clear correlation is seen between high values of the standard deviation of the added Gaussian noise and high density of high anomaly scores. 

\begin{figure}[!t]
\begin{minipage}{1.0\linewidth}
  \centering
  \centerline{\includegraphics[width=8.5cm]{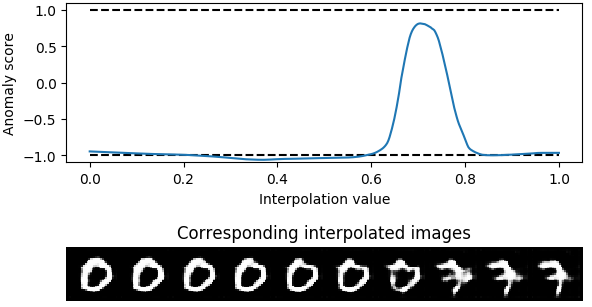}}
\end{minipage}
\caption{Method trained on MNIST and evaluating scores on images generated from interpolated latent variables $(1-t)z_1+tz_2$, for $t\in [0,1]$.}
\label{interp}
\end{figure}
As a sanity check, we take the final state of the generator, trained on MNIST with (\ref{eq:WGAN}) and \cite{WGAN-GP} algorithm, and verify that our method is able to detect generated sample that do not belong to the normal MNIST distribution. In Figure \ref{interp}, we randomly select two latent variable ($z_1$ and $z_2$) which are confidently classified as normal, then linearly interpolate all latent variables between them, given by $(1-t)z_1+tz_2, \forall t \in [0,1]$. Finally, we evaluate the anomaly score of each generated image.

\begin{figure}[!t]
\begin{minipage}{1.0\linewidth}
  \centering
  \centerline{\includegraphics[width=8.5cm,height=4cm]{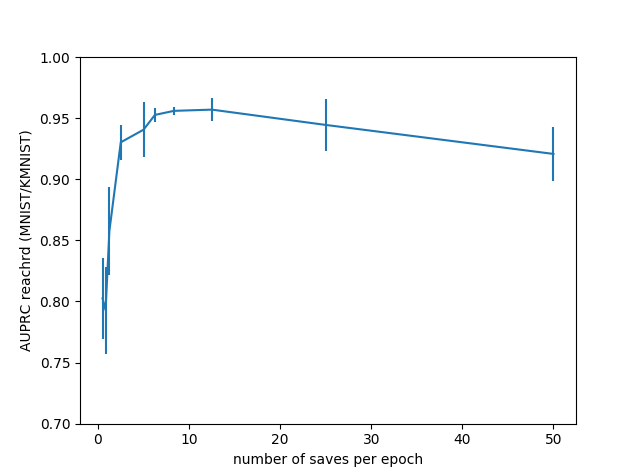}}
\end{minipage}
\caption{Influence of the number of checkpoints per training epoch}
\label{n}
\end{figure}
Finally we check the influence of the number of saves per epoch on the performance of the model.
Figure \ref{n} displays the AUPRC of normal data (MNIST) against KMNIST dataset for different values of saving frequency during WGAN training. For low values, the information carried by $\P{G_\text{hist}}$ starts at a 'early stopping of GANs' (\cite{ssod}) level, and gets richer as the number of saves per epoch increases; hence the increase in AUPRC. We do not have an explanation for the small decay in performance for big values.

Fig.~\ref{fig:scores} compares six state-of-the-art anomaly detection methods with the presented method with both experimental cases and with our $D_{TV}$ initialization (denoted as no training). Apart from a few digits, HistoryAD challenges state of the art anomaly detection methods.
Fig.~\ref{mnistkmnist} shows the anomaly detection results when our method was trained on MNIST dataset and evaluated on KMNIST. Notice that most of the histogram mass of normal and anomalous data is located around -1 and 1, respectively.
This figure 
empirically proves the robustness of the method to anomalous data structurally close to training data.
On the other hand, Fig.~\ref{svhn} shows how well the method performs on structurally different data. Our method was trained on Street View House Numbers \cite{SVHN}, and reached high AUPRC results. Both the approximate density and the AUPRC comparison show that the presented method is able to discriminate anomalous from normal data.

\section{Conclusions and future work}
\label{sec:conclusions}
In this paper, we presented our new anomaly detection approach, HistoryAD, and estimated its performance. Unlike many GAN-based methods, we do not try to invert the generator's mapping, but use the rich information of the whole training process, yielding an efficient and general anomaly detector.
Further can be done in exploiting the training process of GANs, for instance, using multiple training histories to improve the adversarial complexity of $\P{G_\text{hist}}$.


\section*{Appendix}
The goal of this appendix is to obtain a solution of the minimization problem
\begin{equation}
    \min_D \mathcal{L}(D)
\end{equation}
where $\mathcal{L}(D)$ is given by (\ref{mixed objective}). Assuming that the probability distributions $\pd$ and $\pg$ admit densities $d\pd(x)$ and $d\pg(x)$, respectively, the loss can be written as integral of the point-wise loss $\textit{l}$ defined below in (\ref{point-wise objective}):
\begin{equation}
    \mathcal{L}(D)=\int \textit{l}(D(x))dx
\end{equation}
where
\begin{align}
\textit{l}(D(x))=&(d\pd(x) - d\pg(x)) D(x) \nonumber \\&+ \lambda \frac{d\pd(x)+d\pg(x)}{2} d(D(x),[-1,1])^2.
 \label{point-wise objective}
\end{align}
Let us recall that $D_{TV}^{*}=\mathbbm{1}_{P}-\mathbbm{1}_{N}$ where $(P, N)$ is the Hahn decomposition of $d\P{G_\text{hist}}-d\P{\text{data}}$ (therefore, $\textrm{sign}(D_{TV}^{*})=D_{TV}^{*}$).

We notice that for all $x$ and for all $\epsilon>0$,
\begin{align}
\textit{l}[(1-\epsilon)D_{TV}^*(x)]&\ge (d\pd - d\pg)(x)(1-\epsilon)D_{TV}^*(x) \label{dpos}\\
&> (d\pd - d\pg)(x)D_{TV}^*(x) \label{tvdef}\\ \textrm{i.e.}& > \textit{l}[D_{TV}^*(x)] \label{d}
\end{align}  
Indeed, inequality (\ref{dpos}) comes from the positivity of the distance $d(\cdot,[-1,1])$.
On the other hand, inequality (\ref{tvdef}) comes from the definition of $D_{TV}^*$. Indeed, 
if $d\pd(x)-d\pg(x)<0$, then $D_{TV}^*(x)=1$;
the other case $d\pd(x)-d\pg(x)>0$ gives $D_{TV}^*(x)=-1$.
Either way, we obtain that $(d\pd(x) - d\pg(x))(1-\epsilon)D_{TV}^*(x) 
> (d\pd(x) - d\pg(x))D_{TV}^*(x)$.
Finally, inequality (\ref{d}) is obtained from $d(D_{TV}^*(x),[-1,1])=0$.

We can always write a real function $D$ as $D =D_{TV}^*+\Delta$, where $\Delta$ is a certain function. We just proved that if $\textrm{sign}(\Delta(x)) = -\textrm{sign}(D_{TV}^*(x))$ on a non-negligible set, then $D$ cannot minimize (\ref{mixed objective}), since $D_{TV}^*(x)$ achieves lower value than $D(x)$ on this set. 

Hence all minimizer $D^*$ of (\ref{mixed objective}) must be of the form $D^*(x) =(D_{TV}^*+\Delta)(x)$, where $\textrm{sign}(\Delta) = \textrm{sign}(D_{TV}^*)$ almost everywhere.
We can now re-write the point-wise loss formula (\ref{point-wise objective}) as
\begin{align}
\textit{l}(D(x))&=(d\pd(x) - d\pg(x))\cdot (D_{TV}^*(x)+\Delta(x)) \\&+ \lambda \frac{d\pd(x)+d\pg(x)}{2} \Delta(x)^2 (x)\\
&=-2\cdot \delta(\pd,\pg)\\&+\int (d\pd - d\pg)\cdot \Delta + \lambda \frac{d\pd+d\pg}{2} \Delta^2
\end{align}
Minimizing this point-wise second order equation in $\Delta$, we obtain
\begin{equation}
    \Delta^*(x)=\frac{d\pg(x)-d\pd(x)}{\lambda(d\pg(x)+d\pd(x))}.
\end{equation}
Finally, the minimum loss is
\begin{equation}\label{eq:minlossbis}
    -2\cdot \delta(\pd,\pg) -\frac{1}{2\lambda}\int \frac{(d\pg(x)-d\pd(x))^2}{(d\pg(x)+d\pd(x))} dx.
\end{equation}

{\small
\bibliographystyle{ieee_fullname}
\bibliography{egbib}
}

\end{document}